\documentclass[letterpaper]{article} % DO NOT CHANGE THIS
\usepackage{aaai25}  % DO NOT CHANGE THIS
\usepackage{times}  % DO NOT CHANGE THIS
\usepackage{helvet}  % DO NOT CHANGE THIS
\usepackage{courier}  % DO NOT CHANGE THIS
\usepackage[hyphens]{url}  % DO NOT CHANGE THIS
\usepackage{graphicx} % DO NOT CHANGE THIS
\usepackage{natbib}  % DO NOT CHANGE THIS AND DO NOT ADD ANY OPTIONS TO IT
\usepackage{caption} % DO NOT CHANGE THIS AND DO NOT ADD ANY OPTIONS TO IT
\usepackage{algorithm}
\usepackage{algorithmic}

\usepackage{newfloat}
\usepackage{listings}
\DeclareCaptionStyle{ruled}{labelfont=normalfont,labelsep=colon,strut=off} % DO NOT CHANGE THIS
\floatstyle{ruled}
\newfloat{listing}{tb}{lst}{}
\floatname{listing}{Listing}
\usepackage{subcaption}
\usepackage{xcolor}
\definecolor{codegreen}{rgb}{0,0.6,0}
\definecolor{codegray}{rgb}{0.5,0.5,0.5}
\definecolor{codepurple}{rgb}{0.58,0,0.82}
\definecolor{backcolour}{rgb}{0.95,0.95,0.92}

\lstdefinestyle{mystyle}{
    backgroundcolor=\color{backcolour},   
    commentstyle=\color{codegreen},
    keywordstyle=\color{magenta},
    numberstyle=\tiny\color{codegray},
    stringstyle=\color{codepurple},
    basicstyle=\ttfamily,
    breakatwhitespace=false,         
    breaklines=true,                 
    captionpos=b,                    
    keepspaces=true,                 
    numbers=left,                    
    numbersep=5pt,                  
    showspaces=false,                
    showstringspaces=false,
    showtabs=false,                  
    tabsize=2
}

\usepackage{tikz}
\tikzstyle{arrow}=[->] % try "-to", "-stealth", "->", "-latex"
\usetikzlibrary{calc}
\usetikzlibrary{bayesnet}
\usetikzlibrary{shapes.geometric}
\usetikzlibrary{shadows} 
\usetikzlibrary{positioning, shapes.geometric, arrows.meta}
\definecolor{pastel_blue}{HTML}{a9def9}
\definecolor{pastel_purple}{HTML}{cdc1ff}
\definecolor{pastel_green}{HTML}{d3f8e2}
\definecolor{pastel_yellow}{HTML}{ffee93}
\definecolor{pastel_orange}{HTML}{FF9966}
\definecolor{pastel_red}{HTML}{ee6055}
\pgfdeclarelayer{fg}    % declare foreground layer
\pgfdeclarelayer{mid}    % declare middle layer
\pgfdeclarelayer{bg}    % declare background layer
\pgfsetlayers{bg, mid, main, fg}  % set the order of the layers (main is the standard layer)

\usepackage{mathtools}
\usepackage{multirow}
\usepackage{amsmath}
\usepackage{makecell}
\usepackage{amssymb}
\usepackage{amsfonts}
\newcommand{\Assign}{\ensuremath{\leftarrow}}
\newcommand{\tcp}[1]{\texttt{\textcolor{gray}{// #1}}}

\title{MarkovType: A Markov Decision Process Strategy for Non-Invasive Brain-Computer Interfaces Typing Systems}

\author{
    Elifnur Sunger\textsuperscript{\rm 1},
    Yunus Bicer\textsuperscript{\rm 1},
    Deniz Erdogmus\textsuperscript{\rm 1},
    Tales Imbiriba\textsuperscript{\rm 2},
}
\affiliations{
    \textsuperscript{\rm 1}Northeastern University, Boston, MA, USA\\
    \textsuperscript{\rm 2}University of Massachusetts Boston, Boston, MA, USA
    \textsuperscript{\rm 1}\{sunger, bicer, erdogmus\}@ece.neu.edu, \textsuperscript{\rm 2}tales.imbiriba@umb.edu
    
}

\nocopyright

\begin{document}

\maketitle

\begin{abstract}
Brain-Computer Interfaces (BCIs) help people with severe speech and motor disabilities communicate and interact with their environment using neural activity. This work focuses on the Rapid Serial Visual Presentation (RSVP) paradigm of BCIs using noninvasive electroencephalography (EEG). The RSVP typing task is a recursive task with multiple sequences, where users see only a subset of symbols in each sequence. Extensive research has been conducted to improve classification in the RSVP typing task, achieving fast classification. However, these methods struggle to achieve high accuracy and do not consider the typing mechanism in the learning procedure. They apply binary target and non-target classification without including recursive training. To improve performance in the classification of symbols while controlling the classification speed, we incorporate the typing setup into training by proposing a Partially Observable Markov Decision Process (POMDP) approach. To the best of our knowledge, this is the first work to formulate the RSVP typing task as a POMDP for recursive classification. Experiments show that the proposed approach, MarkovType, results in a more accurate typing system compared to competitors. Additionally, our experiments demonstrate that while there is a trade-off between accuracy and speed, MarkovType achieves the optimal balance between these factors compared to other methods.

\end{abstract}

\begin{links}
    \link{Code}{https://github.com/neu-spiral/MarkovType}
    % \link{Datasets}{https://aaai.org/example/datasets}
    % \link{Extended version}{https://aaai.org/example/extended-version}
\end{links}
% \section{Introduction}
\section{Introduction}\label{sec:Introduction}
% \todo{generic EEG classiciation}

Brain-Computer Interfaces (BCIs) allow people with severe speech and motor disabilities to communicate and interact with their environments~\citep{orhan2012rsvp}. Recent BCI research spans applications like typing tasks~\citep{bigdely2008brain, lawhern2018eegnet, smedemark2023recursive} and prosthetic arm control~\citep{bright2016eeg}. Common BCI paradigms include steady-state visual evoked potentials (SSVEP)~\citep{norcia2015steady}, imagined motor movements (MI)~\citep{wierzgala2018most}, rapid serial visual presentation (RSVP)~\citep{acqualagna2010novel}, and matrix layout-based P300-speller~\citep{krusienski2008toward}, which uses P300 event-related potentials observed in EEG to enable individuals to communicate without relying on voluntary muscle activity~\citep{farwell1988talking,mak2011optimizing}. In certain cases, one paradigm may be more suitable than others. For example, individuals with severe speech and motor impairments, such as those with locked-in syndrome, ALS, or cerebral palsy, might not have the precise gaze control required for eye-tracking systems and matrix layout-based P300 spellers. However, precise gaze control is not required in RSVP to distinguish between symbols~\citep{acqualagna2010novel, orhan2012rsvp}. 

% In certain cases, one paradigm may be more suitable than others. For example, people with the locked-in syndrome might not have the precise gaze control needed for eye-tracking systems and the matrix layout-based P300-speller. However, precise gaze control is not required in RSVP to distinguish between symbols~\citep{acqualagna2010novel, orhan2012rsvp}.

This paper focuses on the RSVP paradigm, where users type a target symbol from a predefined alphabet without physical movement, while the system rapidly presents candidate symbols (a query) and collects continuous EEG data. Users perform multiple sequences, and the model updates symbol probabilities based on their responses to each query. Typing stops when the probability of any symbol exceeds a confidence threshold or when the maximum number of sequences is reached. In RSVP setups, users do not observe the entire set of symbols at once. Instead, they observe a subset of symbols at each sequence. This makes RSVP suitable for formulation as a partially observable Markov decision process (POMDP). In POMDPs, the true state of the environment is unknown, and the agent or neural network receives only partial observations at each step~\citep{kaelbling1998planning}.

We introduce ``MarkovType'', which formulates the RSVP typing task as a POMDP and enhances the recursive typing performance. We state our contributions as follows:
\begin{itemize}
    \item We propose a Markov Decision Process for non-invasive BCI typing systems (MarkovType) and formulate the BCI typing procedure as a Partially Observable Markov Decision Process (POMDP), incorporating the typing mechanism into the learning procedure. To the best of our knowledge, this is the first time that the BCI typing procedure has been formalized as a POMDP and used for the typing task classification.
    \item We evaluate MarkovType in a simulated typing task and show it outperforms previous approaches using Recursive Bayesian Estimation in classification accuracy and information transfer rate.
    \item We also show that there is a trade-off between accuracy and the number of sequences in the typing task. This opens the door for future improvements and explorations.
\end{itemize}

% This paper is organized as follows: Section~\ref{sec:RelatedWork} discusses current research on RSVP typing tasks and applications of POMDP. In Section~\ref{sec:methods}, we define the problem addressed in this work and introduce MarkovType. Section~\ref{sec:experiments} covers the dataset and model details, competing methods, experimental procedures, and evaluation metrics. In Section~\ref{sec:results}, we present the performance of MarkovType compared to competing methods and discuss its limitations. Finally, Section~\ref{sec:conclusion} summarizes MarkovType, our results, and contributions.

% \section{Related Work}
\section{Related Work}\label{sec:RelatedWork}
% \subsection{RSVP}
Extensive research exists on RSVP classification using EEG data. Prior work on classification has focused on using logistic regression or linear discriminant analysis~\citep{10.1371/journal.pone.0044464, bigdely2008brain}, and applying the Recursive Bayesian rule for classification~\citep{higger2013robust}. In addition to classification, Recursive Bayesian methods are used for query optimization~\citep{koccanaougullari2018analysis, koccanaougullari2019active, marghi2022active}, and designing stopping criteria for typing tasks~\citep{koccanaougullari2021stopping}. 

There are also prior works on using convolutional neural networks (CNNs) for single-trial EEG RSVP classification~\citep{shamwell2016single, zang2021deep, lawhern2018eegnet}. However, these methods cannot perform recursive classification. \citet{smedemark2023recursive} focuses on a recursive Bayesian classification based on a discriminative probability for each query, which they approximate using neural network classifiers such as EEGNet~\cite{lawhern2018eegnet} and hand-crafted CNNs. 
They show that discriminative models using hand-crafted CNNs outperform generative models such as linear discriminant analysis and logistic regression. They do not include the typing procedure in training and assume responses within a query are conditionally independent. Consequently, they perform binary classification, predicting each response in a single query as either a ``target'' or ``non-target''. Although the proposed method performs well in binary classification, symbol classification can be improved by including the typing task in training and reconsidering the independence assumption to allow classification over the alphabet. We propose to do this by formulating the RSVP typing task as a partially observable Markov decision process (POMDP).
% -classifiers have been implemented for
% classification of RSVP tasks

% LDA

% robust,detection of target images in a rapid serial visual presentation (RSVP) task based on EEG data 
% \citet{10.1371/journal.pone.0044464}
% ICA \citet{bigdely2008brain}

% -Recursive 
% Recursive Bayesian for SSVEP \citet{higger2016recursive}

% Classification rule that is robust to changes in the assumed data distributions, tested on the RSVP (using bayes rule for the fusion model with using RDAfor feature extraction) \citet{higger2013robust}

% -query selection/optimization:

% an active querying procedure using mutual information maximization in recursive state estimation \citet{koccanaougullari2018analysis}

% a novel recursive state estimation framework
% for BCI-based typing systems using active querying through sequences of stimuli
% and stopping criterion \citet{koccanaougullari2019active}

% \citet{marghi2022active}

% -Stopping criteria
% the analysis and design of
% stopping criterion for recursive Bayesian classification \citet{koccanaougullari2021stopping}

% -CNN

% Single-trial EEG RSVP classifier \citet{shamwell2016single}
% single trial \citet{zang2021deep}
%  EEGNet, a hand-designed 2D convolutional neural network (CNN), again single trial \citet{lawhern2018eegnet}

% -
% Recursive bayesian for simulated typing task on simulated data, build upon a variety
% of recent work performing Bayesian updates to symbol probabilities
% that largely use generative models for computing updates \citet{smedemark2023recursive}

% \subsection{POMDP}
POMDP is used in many applications, such as visual attention networks~\citep{mnih2014recurrent}, human-robot interaction~\citep{chen2018planning}, autonomous driving~\citep{bai2015intention}, and medical diagnosis~\citep{ayer2012or, hauskrecht2000planning}. In the BCI domain, numerous works use POMDP-based models for symbol classification. \citet{park2012pomdp, park2010pomdp} employed a POMDP-based model for classification using a matrix speller, while~\citet{tresols2023pomdp} proposed a POMDP-based model for high-level decision making across three BCI modalities: SSVEP~\cite{norcia2015steady}, MI~\citep{wierzgala2018most}, and code-modulated visual evoked potentials (CVEP)~\citep{sutter1992brain}. However, to the best of our knowledge, no prior work formulate the RSVP typing task as a POMDP.

% - POMDP examples ...

% - POMDP used for image classification with attention: recurrent models of visual attention \citet{mnih2014recurrent}

% - POMDP in BCI 
% matrix speller \citet{park2012pomdp, park2010pomdp}

% a model based on POMDP that works as a high-level decision making framework for three different active/reactive BCI modalities: steady-state visual evoked potentials (SSVEP)~\cite{norcia2015steady}, imagined motor movements (MI)~\citep{wierzgala2018most}, CVEP~\citep{sutter1992brain} \citet{tresols2023pomdp}

% \section{Methods}
\section{Methods}\label{sec:methods}
\subsection{Problem Statement}
\label{sec:ProblemStatement}

\begin{figure}[t]
\centering
\begin{tikzpicture}[
    node distance=0.45cm and 0.45cm,
    every node/.style={draw, minimum size=0.45cm},
    arrow/.style={-{Latex[scale=1]}, thick, }
]

% Define nodes
% \node (data) [circle] {\( E \)};
% Define nodes
\node (label) [circle] {\( t \)};
\node (pnm1) [below= of label, rectangle, rounded corners] {\( \mathbf{p}_{n\!-\!1} \)};
\node (simulator) [right= 0.75 cm of label, rectangle, dashed] {Simulator};
\node (q) [right = of simulator, circle] {\( \mathbf{q}_n \)};
\node (Dq) [below= of q, circle] {\( \mathbf{E}_n \)};
\node (fe) [ right= of q, rectangle] {\( f_{e}(\boldsymbol\theta_{e}) \)};
\node (gn) [below=of fe, rectangle, rounded corners] {\( \mathbf{G}_n \)};
\node (fh) [below=of gn, rectangle] {\( f_{h}(\boldsymbol\theta_{h}) \)};
\node (hnm1) [left= 3.9cm of fh, rectangle, rounded corners] {\( \mathbf{h}_{n\!-\!1} \)};
\node (hn) [right=of fh, rectangle, rounded corners] {\( \mathbf{h}_n \)};
\node (fc) [above=of hn, rectangle] {\( f_{c}(\boldsymbol\theta_{c}) \)};
\node (pn) [above=0.4cm of fc, rectangle, rounded corners] {\( \mathbf{p}_n \)};

% Draw arrows
% \draw [arrow] (data) -- (simulator);
\draw [arrow] (label) -- (simulator);
\draw [arrow] (pnm1) -- (simulator);
% \draw [arrow] (simulator) -- (Dq);
\draw [arrow] (simulator) -- (q);
\draw [arrow] (q) -- (Dq);
\draw [arrow] (Dq) -- (fe);
\draw [arrow] (q) -- (fe);
\draw [arrow] (fe) -- (gn);
\draw [arrow] (gn) -- (fh);
\draw [arrow] (hnm1) -- (fh);
\draw [arrow] (fh) -- (hn);
\draw [arrow] (hn) -- (fc);
\draw [arrow] (fc) -- (pn);
% New arrow from pt to some location
\draw [arrow,dashed] (pn) -| +(0.9,0) -| ++(0.9,-0.95) -- ++(0.4,0);
\draw [arrow,dashed] (label) |- +(0,0.55) -| ++(7.4,0.55) -| ++(0,-0.53) -- ++(0.4,0);
\draw [arrow] (hn) -- +(1.3,0);
% Draw dashed outline
\draw [thick, line width=2pt,rounded corners] ([shift={(-0.4cm,0.7cm)}] simulator.south west) rectangle ([shift={(0.38cm,-0.72cm)}] hn.north east);
\node [right=1.0cm, yshift=-2.25cm, draw=none ] {\textbf{\textit{n=1:N}}};

\end{tikzpicture}
\caption{MarkovType Model Architecture. It consists of a simulator, a feature extractor $f_{e}$, a core network $f_{h}$ and a classification network $f_{c}$. A single typing trial includes  $N$ sequences. Given the target symbol $t$ and the prior over the alphabet $\mathbf{p}_{n\!-\!1}\in\mathbb{R}^{A}$, the simulator extracts a query with $K$ unique symbols $\mathbf{q}_n\in \mathbb{Z}^{K}$ and EEG responses $\mathbf{E}_{n}\in \mathbb{R}^{K\times c\times f}$, where $A$ is the length of the alphabet, $c$ is the number of EEG channels and $f$ is the feature length. $f_e$ extracts features  from $\mathbf{E}_{n}$ and maps them over the alphabet using $\mathbf{q}_{n}$ as in Eq.~\eqref{eq:feature_mapping} to obtain $\mathbf{G}_{n} \in \mathbb{R}^{A\times L}$, where $L$ is the feature length. $f_{h}$ forms the hidden states of the current sequence $\mathbf{h}_{n}$ using the alphabet features $\mathbf{G}_{n}$ and the hidden states of the previous sequence ($\mathbf{h}_{n\!-\!1}$), which provides knowledge of the responses to previous queries.}
\label{fig:architecture}
\end{figure}
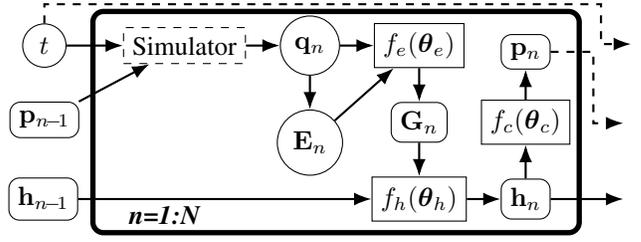
This work focuses on an RSVP typing task with $N$ independent sequences, where the user tries to type the symbol $t$ from an alphabet of $A$ letters. At each sequence, the user sees a query of symbols indexed by $\mathbf{q}_{n}\in \mathbb{Z}^{K}$ with $K$ different symbols, where $K\leq A$ and the index of target letter $t$ may or may not be in $\mathbf{q}_{n}$. We observe the EEG responses to the queried symbols, $\mathbf{E}_{n} \in \mathbb{R}^{K\times c\times f}$, where $c$ is the number of EEG channels and $f$ is the feature-length. Note that the user never sees all the symbols in the alphabet simultaneously; they only see a subset of symbols appearing sequentially on the screen within a sequence.
% Note that the user never sees all symbols in the alphabet simultaneously. 

To perform classification on this typing task, previous works assume that the EEG responses $\mathbf{E}_{n}$ to the $K$ symbols in the query $q_n$ are independent~\citep{smedemark2023recursive, koccanaougullari2018analysis}.
As a result, they classify each symbol in the query as either ``target'' or ``non-target'', without incorporating the recursive typing task into training.

Here, we include the RSVP typing task in the training procedure, where the model has partial observations (never receiving responses for all symbols in the alphabet simultaneously) and jointly learns EEG feature extraction and a fusion process for recursive classification over multiple sequences. To achieve recursive classification from partial observations and to incorporate the typing process in both training and testing, we formulate the typing process as a POMDP.

We propose a recursive classification model for the RSVP typing task, inspired by the recursive classification network of the Recurrent Attention Model (RAM) proposed by~\citet{mnih2014recurrent}. In our setup, as illustrated in Figure~\ref{fig:architecture}, the agent (which is the model itself) learns to extract features from a query and incorporates hidden states recursively to perform classification over the alphabet $A$, while receiving a scalar reward that it aims to maximize. During this process, the agent chooses a query $\mathbf{q}_n$ from the prior over the alphabet $\mathbf{p}_{n-1}$, which is updated with each sequence.

At the first sequence, we assume a uniform prior over the alphabet $\mathbf{p}_{0}\in\mathbb{R}^{A}$, which is then updated after each sequence. Note that non-uniform priors can be incorporated, often employing language models~\citep{Orhan_2013, e12b0b4cff2140fa9507b111ce7595a9}. However, we opted to analyze the performance of MarkovType without the influence of language models. Details on the prior update are given in Section~\ref{sec:ModelArchitecture}. 

% Following the RSVP paradigm, we focus on recursively classifying typing sequences while incorporating previous features, extracted using $\mathbf{E}_{n}$ and $\mathbf{q}_{n}$. In the RSVP typing setup, the network never fully observes the environment, meaning it never receives responses for all symbols in the alphabet simultaneously. To achieve recursive classification from partial observations and to include the typing process in both training and testing, we formulate the typing process as a POMDP. To cope with the recursive nature of RSVP typing systems and to achieve recursive classification within a POMDP setup, we are inspired by the recurrent classification network of the Recurrent Attention Model (RAM) proposed by~\citet{mnih2014recurrent}, which introduces a recurrent neural network model for image classification with a POMDP setup coming from its attention mechanism. We propose a recursive classification model for the RSVP typing task that includes a simulator to sample the query from the alphabet using a prior over the alphabet, which is updated at each sequence. 

% At each sequence, the network receives a scalar reward that it aims to maximize.

\subsection{MarkovType Model Architecture}
\label{sec:ModelArchitecture}
The overall model architecture, as shown in Figure~\ref{fig:architecture}, includes a simulator, a feature extractor $f_{e}$, followed by a core network $f_{h}$ and a classification network $f_{c}$. In this setup, a single typing trial consists of $N$ sequences. Given the target symbol $t$ and the prior over the alphabet $\mathbf{p}_{n\!-\!1}$, the simulator extracts the query $\mathbf{q}_n$ and EEG responses $\mathbf{E}_{n}$. $f_e$ extracts features $\mathbf{G}_{n} \in \mathbb{R}^{A\times L}$ from $\mathbf{E}_{n}$ and maps it over the alphabet using $\mathbf{q}_{n}$ as in Eq.~\eqref{eq:feature_mapping}, where $L$ is the feature length. 
\begin{equation}
    \label{eq:feature_mapping}
    \mathbf{G}_{n,i} = 
    \begin{cases} 
    f_e(\mathbf{E}_{n,i}) & \text{if } i \in \mathbf{q}_{n} \, , \\
    \mathbf{0}_L & \text{otherwise} \, .
    \end{cases}
\end{equation}
$\mathbf{0}_L$ is a zero array with length $L$, $\mathbf{E}_{n,i}$ is the EEG response, and $\mathbf{G}_{n,i}$ are the features for the symbol $i \in A$ at sequence $n$.
The hidden states of the current sequence $\mathbf{h}_{n} \in \mathbb{R}^{v}$ (where $v$ is the length of the hidden feature) are formed by $f_{h}$ using the alphabet features $\mathbf{G}_{n}$ and the hidden states of the previous sequence ($\mathbf{h}_{n\!-\!1}$), which provide information on previous responses. Please see Section~\ref{sec:dataset and model details} for model details. 

% the simulator, feature extractor, core network and classification network.

\subsubsection{Actions.}
At each sequence, the agent (neural network) performs an environment action $\mathbf{q}_{n} \in \mathbb{Z}^{K}$, where each element of $\mathbf{q}_{n}$ is sampled from $\mathbf{p}_n = f_c(\mathbf{h}_n;\boldsymbol\theta_c)$. This action affects the next state of the environment ($ \mathbf{E}_{n+1}$). 
% $\mathbf{p}_n$ affects the reward that the neural network receives. \todo{todo}
% This is why 
% target classes are $A$ alphabet symbols, rather than target or non-target binary classifications in most works~\citet{lawhern2018eegnet, smedemark2023recursive, koccanaougullari2018analysis}.
\subsubsection{Reward.}
The setup explained in Sections~\ref{sec:ProblemStatement} and \ref{sec:ModelArchitecture} is a particular case of the POMDP. In the POMDP setup for this problem, the user's EEG responses to all symbols in the alphabet are unknown and the agent (in our case, the neural network) receives partially observed sets of responses $\mathbf{E}_n$ to the query $\mathbf{q}_n$ at each sequence.

The total reward function according to POMDP is given in Eq.~\eqref{eq:POMDP reward}, where $r_{n}$ is the reward assigned at sequence $n$. 
\begin{equation}
    \label{eq:POMDP reward}
    R=\sum_{n=1}^{N}r_{n} d(n)
\end{equation}
$r_{n}$ is equal to 1 if the predicted label $\hat{t}=\arg\max_{i\in A}(\mathbf{p}_{n,i})$ matches the correct target label $t$ at sequence $n$, and 0 otherwise, where $\mathbf{p}_{n,i}$ is the probability of symbol $i$. $d(n)$ is the discount factor. It creates a trade-off between finding the actions with the largest immediate reward and those with the largest sum of rewards. See Section~\ref{sec:implementation_details} for details on $d(n)$.

\subsection{Learning Rule}
\label{learning rule}
The neural network needs to learn a stochastic policy $\pi(( \mathbf{q}_{n})|\mathbf{s}_{1:n}; \boldsymbol\theta)$  which maps the history of past interactions with the environment to a distribution over actions for the current sequence. The history of past interactions $\mathbf{s}_{1:n} = \mathbf{E}_{1}, \mathbf{q}_{1}, . . ., \mathbf{E}_{n\!-\!1}, \mathbf{q}_{n\!-\!1},  \mathbf{q}_{n}, \mathbf{E}_{n}$ where $\mathbf{E}_{n}$ is the observation produced by the environment, and $\mathbf{q}_{n}$ are the actions. In this paper's setup,  the policy $\pi$ is defined by the model, and the history $\mathbf{s}_{n}$ is summarized in the hidden state $\mathbf{h}_{n}$.

Model parameters are learned by maximizing the total reward the model expects when interacting with the environment, using the REINFORCE rule~\citep{wierstra2007solving}.

\subsubsection{POMDP - REINFORCE RULE.}

The REINFORCE rule~\citep{wierstra2007solving} follows the policy gradient framework, where model weights $\boldsymbol\theta$ are updated directly by estimating the gradient in the direction of higher reward. $\mathbf{s}_{1:n}$ is the observed history, but we also need to define the complete history $\mathbf{H}$, which includes the unobserved (future) states. It is expressed by $\mathbf{H}_{N} = \langle \mathbf{s}_{N}, \mathbf{G}_{1:N} \rangle$. The probability of a history for given policy-defining weights $\boldsymbol\theta$ is $p(\mathbf{H}|\boldsymbol\theta)$, and $R(\mathbf{H})$ is a measure of the total reward achieved during history $\mathbf{H}$, which we aim to maximize in Eq.~\eqref{trainingaim}.

\begin{equation}
    \label{trainingaim}
    J(\boldsymbol\theta) = \int_{H}p(\mathbf{H}|\boldsymbol\theta)R(\mathbf{H})\mathrm{d}\mathbf{H} \, .
\end{equation}
We rewrite Eq.~\eqref{trainingaim} using gradient ascent to update $\boldsymbol\theta$:
\begin{equation}
    \label{gradient}
    \nabla_{\boldsymbol\theta}J=\int\nabla_{\boldsymbol\theta} p(\mathbf{H}|\boldsymbol\theta)R(\mathbf{H})\mathrm{d}\mathbf{H}  \, .
\end{equation}
Viewing the problem as a POMDP setup, we can first apply the ``likelihood-ratio trick'' (also considering the fact that $\nabla_{\boldsymbol\theta}R=0$ for a single, fixed $\mathbf{H}$), as follows:

\begin{equation}
\begin{aligned}
    \label{likelihoodtrick}
    \nabla_{\boldsymbol\theta}J &= 
    \int\frac{p(\mathbf{H}|\boldsymbol\theta)}{p(\mathbf{H}|\boldsymbol\theta)}\nabla_{\boldsymbol\theta} (p(\mathbf{H}|\boldsymbol\theta))R(\mathbf{H})\mathrm{d}\mathbf{H} \, ,\\
    &= \int p(\mathbf{H}|\boldsymbol\theta)\nabla_{\boldsymbol\theta} \log(p(\mathbf{H}|\boldsymbol\theta))R(\mathbf{H})\mathrm{d}\mathbf{H} \, .
\end{aligned} 
\end{equation}
We take the sample average with the Monte Carlo approximation (MC) in Eq~\eqref{eq:MC sampling}, where M is the number of episodes.
\begin{equation}
\begin{aligned}
    \label{eq:MC sampling}
    \nabla_{\boldsymbol\theta}J &=\mathbb{E}_{H}[\nabla_{\boldsymbol\theta} \log(p(\mathbf{H}|\boldsymbol\theta))R(\mathbf{H})] \\
     &\approx\frac{1}{M}\sum_{m=1}^{M}\nabla_{\boldsymbol\theta} \log(p(\mathbf{H}^{m}|\boldsymbol\theta))R(\mathbf{H}^{m}) \, .
\end{aligned}
\end{equation}
To calculate $\log p(\mathbf{H}|\boldsymbol\theta)$, we state that the probability of a particular history is the product of all actions and observations given the sub-histories as follows:

% \begin{equation}
% \begin{aligned}
\begin{align}
    \label{pht}    p(\mathbf{H}_{N}|\boldsymbol\theta)\!=\!p(\mathbf{E}_{1},\mathbf{G}_{1})\prod_{t=2}^{N}&\big[p(\mathbf{E}_{n},\mathbf{G}_{n}|\mathbf{s}_{1:n\!-\!1},\mathbf{q}_{n\!-\!1},\mathbf{G}_{1:n}) \nonumber\\
    &\times\pi(\mathbf{q}_{n\!-\!1}|\mathbf{s}_{1:n\!-\!1};\boldsymbol\theta) \big]\, ,
\end{align}
% \end{aligned}
% \end{equation}
% where $\mathbf{u}_{n}=\mathbf{p}_{n}$ and $\mathbf{E}_{n} = (\mathbf{E}_{n},\mathbf{q}_{n})$. Then, taking the log-derivate of Eq. (\ref{pht}) results in:
Then, taking the log-derivate of Eq. (\ref{pht}) results in:
\begin{equation}
    \label{logderiv_pht}
    \nabla_{\boldsymbol\theta} \log p(\mathbf{H}_{N}|\boldsymbol\theta) = \sum_{n=1}^{N}\nabla_{\boldsymbol\theta}\log \pi(\mathbf{q}_{n}|\mathbf{s}_{1:n};\boldsymbol\theta) \, .
\end{equation}
Therefore, the gradient approximation can be written as:
\begin{equation}
    \label{grad app}
    \nabla_{\boldsymbol\theta}J
    \approx\frac{1}{M}\sum_{m=1}^{M}\sum_{n=1}^{N}\nabla_{\boldsymbol\theta}\log \pi(\mathbf{q}_{n}^{m}|\mathbf{s}_{1:n}^{m};\boldsymbol\theta)R_{n}^{m} \, ,
\end{equation}
where $R_n^{m}$ is the sum of rewards at sequece $n$ and episode $m$. Note that $\nabla_{\boldsymbol\theta}\log \pi(\mathbf{q}_{n}^{m}|\mathbf{s}_{1:n}^{m};\boldsymbol\theta)$ is the gradient of the model and can be computed by standard backpropagation. Also, to overcome the high variance problem of the gradient due to MC, a baseline \emph{b} is added (that may depend on $\mathbf{s}_{1:n}^{m}$ but not on the actions $\mathbf{q}_{n}^{m}$) as explained by~\citet{mnih2014recurrent}. % Section~\ref{sec:dataset and model details} includes details on our baseline implementation. 
Therefore, we rewrote Eq.~\eqref{grad app} as:

\begin{equation}
    \label{variance red}
    \nabla_{\boldsymbol\theta}J
    \approx\frac{1}{M}\sum_{m=1}^{M}\sum_{n=1}^{N}\nabla_{\boldsymbol\theta}\log \pi(\mathbf{q}_{n}^{m}|\mathbf{s}_{1:n}^{m};\boldsymbol\theta)(R_{n}^{m}-b_{n}) \, .
\end{equation}

\subsubsection{Hybrid Supervised Loss.}
We follow the supervised loss described by~\citet{mnih2014recurrent}, as defined in Eq.~\eqref{eq:hybrid-supervised loss}.
\begin{subequations}
    \label{eq:hybrid-supervised loss}
    \begin{align}
    \mathcal{L}_{\text{action}} &\!=\! -\log(\mathbf{p}_{N,t}) \, ,\\
    \mathcal{L}_{\text{baseline}} &\!=\!  \frac{1}{N} \sum_{n=1}^{N} (R_{n} - b_{n})^2, \\
    \mathcal{L}_{\text{reinforce}} &\!=\! \!-\frac{1}{M}\sum_{m=1}^{M}\sum_{n=1}^{N}\log \pi(\mathbf{q}_{n}^{m}|\mathbf{s}_{1:n}^{m};\boldsymbol\theta)(R_{n}^{m}\!-\!b_{n}) \, , \\
   \mathcal{L}&\!=\! \mathcal{L}_{\text{action}}+ \lambda\left(\mathcal{L}_{\text{baseline}}+\mathcal{L}_{\text{reinforce}}\right) \, .
    \end{align}
\end{subequations}

\noindent
$\mathcal{L}_{\text{action}}$ is the negative log-likelihood loss, where $\mathbf{p}_{N,t}$ is the predicted probability of the target $t$ at the last sequence, and $\lambda \geq 0$ is a regularization parameter. $\mathcal{L}_{\text{baseline}}$ and $\mathcal{L}_{\text{reinforce}}$ increase the reward and reduce bias in gradient estimation.

% \section{Experiments}
\section{Experiments} \label{sec:experiments}
\subsection{Dataset and Model Details}
\label{sec:dataset and model details}
We perform all experiments on a large, publicly available RSVP benchmark dataset containing over 1 million stimulus presentations~\citep{10.3389/fnins.2020.568000} that is suitable for a simulated typing task. In this study, 64 healthy subjects performed a target image detection task involving street-view images, either with humans (``target'') or without (``non-target''), during 64-channel electroencephalogram (EEG) data recording. We follow the preprocessing steps outlined by~\citet{smedemark2023recursive}. We perform all experiments using data pooled from all subjects, repeating each experiment across 5 randomized train/test splits (with seeds from 0 to 4) and using $10\%$ training set for validation, as described by~\citet{smedemark2023recursive}. 
Details of the simulator, feature extractor $f_e$, the core network $f_h$, baseline $b_n$ and classification network $f_c$ are provided below.
\subsubsection{Simulator.} One can use any type of typing simulator or language model. We use the simulator from \citet{smedemark2023recursive}. This simulator randomly selects $K$ unique symbols from the alphabet based on the prior over the alphabet $\mathbf{p}_{n-1}$ and samples the query of symbols indexed $\mathbf{q}_{n}$. This affects the state of the environment at sequence $n$, i.e., the EEG responses $\mathbf{E}_{n}$ to the queried symbols.

\subsubsection{Feature Extractor.} For feature extraction, we use a hand-designed one-dimensional CNN with 5 convolutional layers, applying 1D convolutions only across the time axis. 
% This is similar to the 1D CNN architecture given by \citet{smedemark2023recursive}.

\subsubsection{Core Network.} The core network $f_h$ is used to update $\mathbf{h}_{n}=$\emph{$LayerNorm(Rect(Linear(\mathbf{h}_{n\!-\!1})+Linear(\mathbf{G}_{n})))$}, where $LayerNorm$ is a normalization layer, $Rect$ is a rectified linear unit layer, and $Linear$ is a linear layer.

% \todo{The core network $f_h$ is used to update $\mathbf{h}_{n}=$\emph{$L_{N}(R(L(\mathbf{h}_{n\!-\!1})+L(\mathbf{G}_{n})))$}, where $L_{N}$ is a normalization layer, $R$ is a rectified linear unit layer, and $L$ is a linear layer.}

\subsubsection{Baseline.} We follow the baseline calculation from the image classification example given by~\citet{mnih2014recurrent} and calculate the baseline at $n$ as $b_{n}=$\emph{$Linear(\mathbf{h}_{n})$}.

\subsubsection{Classification Network.} It is formulated using a softmax output: $f_{c}(\mathbf{h}_{n})=$\emph{Softmax(Linear($\mathbf{h}_{n}$))}. $\mathbf{p}_{{0}}$ is uniformly initialized, and $\mathbf{h}_{0}$ is initialized as an all-zero feature vector.

\subsection{Implementation Details.} \label{sec:implementation_details} We ran experiments on 2.4 GHz Intel E5-2680 v4 CPUs and 2.1 GHz Intel Xeon Platinum 8176 CPUs, using PyTorch~\citep{paszke2019pytorch} 2.3.0. We train the models using the Adam optimizer~\citep{kingma2014adam} with a learning rate of $0.001$ for $200$ epochs, decaying the learning rate by a factor of $0.97$ after each epoch, and use a batch size of $28$. We set the number of symbols shown in a query $K$ to $10$, the number of sequences $N$ to $10$, the number of episodes $M$ to $1$ and the alphabet length $A$ to $28$. We perform hyper-parameter tuning on a validation set consisting of a held-out $10\%$ of the training set, checking $\lambda$ values of $\{0.01, 0.02, 0.03,..., 0.1\}$. 

Our analyses include different discount factors of: $d(n) = \{\frac{2N-n-1}{N},  \frac{1}{n}, \frac{1}{n^2}, \frac{1}{n^3}\}$. These vary in how much importance they give to early correct classifications. Specifically, the importance increases in the given order, from $\frac{2N-n-1}{N}$ to $\frac{1}{n^3}$. The final $\lambda$ values used for models with these discount factors are $\{0.02, 0.02, 0.01, 0.1\}$, respectively. 
% Our analyses include different discount factors of: $d(n) = \{\frac{2N-n-1}{N},  \frac{1}{n}, \frac{1}{n^2}, \frac{1}{n^3}\}$. Note that these discount factors vary in how much importance they give to early correct classifications. Specifically, the importance increases in the given order, from $\frac{2N-n-1}{N}$ to $\frac{1}{n^3}$. The final $\lambda$ values used for models with these discount factors are $\{0.02, 0.02, 0.01, 0.1\}$, respectively. 

\subsection{Competing Methods}
We compare our model against the two best-performing discriminative methods described by~\citet{smedemark2023recursive}, which are 1D and 2D CNNs. Similarly, these methods are trained with the Adam optimizer~\citep{kingma2014adam} using a learning rate of 0.001 for 25 epochs, decaying the learning rate by a factor of 0.97 after each epoch, as described by~\citet{smedemark2023recursive}. We refer to these methods as ``RB - 1D CNN'' and ``RB - 2D CNN'' throughout this paper. Note that during classification, RB methods do not classify $\mathbf{E}_n$ directly over the alphabet (number of A classes). Instead, they classify symbols as ``target'' or ``non-target''. During testing, they receive queries of length $K$, perform binary classification using the Recursive Bayesian rule and map the results to the alphabet, explained by~\citet{smedemark2023recursive}.

\subsection{Experimental Setup}
Algorithm~\ref{alg:algorithm} outlines the testing procedure for the simulated RSVP task using MarkovType\footnote{https://github.com/neu-spiral/MarkovType}. The threshold $\tau$ is used for early stopping. If any posterior probability exceeds this threshold or if the maximum number of sequences (set to 10) is reached, the classification sequence for target $t$ is terminated. Since we are simulating the typing process, we generate $\mathbf{E}_{n}$ by sampling from the ``target'' and ``non-target'' examples given the query $\mathbf{q}_n$ and the symbol $t$. For RB methods, a recurrent Bayesian update occurs instead of line 8 of Algorithm~\ref{alg:algorithm}, which does not use $\mathbf{h}_{n-1}$ as inputs. See Section~\ref{sec:metrics} for Information Transfer Rate (ITR) calculations.
\begin{algorithm}[tb]
\caption{Test procedure}\label{alg:test}
\label{alg:algorithm}
\textbf{Input}: Trained model $f(\cdot)$, 
Pos. and Neg. Test Data $\mathcal{E}^+, \mathcal{E}^-$,
Symbols per query $K$,
Sequences per symbol $N$,
Alphabet size $A$,
Decision threshold $\tau$.\\
\textbf{Output}: ITR (bits per symbol), ITR$_{n_{\tau}}$ (bits per sequence)
\begin{algorithmic}[1] %[1] enables line numbers
\STATE  $C$ \Assign $0$, $N_{\tau}$ \Assign $0$  \tcp*{Correct symbol count, Total number of sequences taken}
\FOR{$t \Assign 1:T$ \tcp*{Target symbols}}
\STATE $\mathbf{h}_{0} \Assign (0,\dots,0)$ \tcp*{Hidden feature initialization}
\STATE $\mathbf{p}_{0} \Assign (\frac{1}{A},\dots,\frac{1}{A})$ \tcp*{Uniform symbol prior}
\FOR{$n \Assign 1:N$ \tcp*{Number of sequences}}
\STATE $\mathbf{q}_{n} \sim \mathbf{p}_{n\!-\!1}$ \tcp*{Sample query}
\STATE $\{ \mathbf{E}_{n}[k]\sim \mathcal{E}^+ \ \textbf{if} \ \mathbf{q}_{n}[k] = t \ \textbf{else} \ \mathbf{E}_{n}[k]\sim \mathcal{E}^- \}_{k = 1}^K$
\STATE $\mathbf{p}_{n},\mathbf{h}_{n} \Assign f(\mathbf{E}_{n},\mathbf{q}_{n}, \mathbf{h}_{n-1})$ \tcp*{Model outputs}
\STATE ind, val $\Assign \arg\max (\mathbf{p}_{n}), \max (\mathbf{p}_{n})$
\IF {val $\geq \tau$ \OR $n=N$}
\IF{ind $=t$}
\STATE $C \Assign C+1$
\ENDIF
\STATE $N_{\tau}$  \Assign $N_{\tau} + n $
\STATE break
\ENDIF
\ENDFOR
\ENDFOR
\STATE  $n_{\tau}$ \Assign $N_{\tau}/T$ \tcp*{Mean number of sequences}
\STATE \textbf{return} ITR($A, C/T$), ITR$(A,C/T)_{n_{\tau}}$
\end{algorithmic}
\end{algorithm}

Our analyses of test time include classifications both with and without a threshold $\tau = 0.8$, using $1000$ target symbols from an alphabet of length $28$.

\subsection{Evaluation Metrics}
\label{sec:metrics}
We compare the accuracy and information transfer rate (ITR)~\citep{shannon1948mathematical} of different models. ITR per selection~\citep{smedemark2023recursive} is calculated as:
\begin{equation}
    ITR(A,P)\!\coloneqq\!\log_{2}(A)\!+\! P\log_{2}(P)\!+\!(1-P)\log_{2}(\frac{1-P}{A-1}),
\end{equation}
where $P$ is the accuracy and $A$ is the alphabet length. However, this formulation does not account for the time to type a symbol. We use ITR per sequence given in Eq.~\eqref{eq:itr_per_attempt} to include the time spent typing a symbol, where $n_{\tau}$ is the mean number of sequences taken for making a decision.
\begin{equation}
    \label{eq:itr_per_attempt}
    ITR(A,P)_{n_{\tau}} = \frac{1}{n_{\tau}}ITR(A,P) \, .
\end{equation}
% \begin{equation}
%     \frac{ITR(A,P)}{\frac{{\sum_{t=1}^{T}N_t}}{T}}, \quad P=\frac{C}{T}
% \end{equation}
% \begin{equation} 
%     ITR(A,P_{attempts}), \quad P_{attempts}=\frac{C}{{\sum_{t=1}^{T}N_t}}
% \end{equation}

% \section{Results}
\section{Results}\label{sec:results}
We evaluate the performance of MarkovType and RB methods both with and without using the threshold $\tau$. When the threshold $\tau$ is applied, the classification sequence for a symbol terminates if any posterior probability exceeds $\tau$ or the maximum number of sequences is reached. This allows us to assess the time spent making a decision (i.e., the number of sequences taken), as well as accuracy and ITR. Without the threshold, early stopping does not occur, and we evaluate performance as a function of the number of sequences, with classification occurring at every sequence. This also demonstrates the upper limit for classification performance.
\subsection{With Threshold $\tau$}
Here, we report the model performances when the testing process for the simulated typing uses the threshold $\tau$ for early stopping. Table~\ref{tab:results} includes a comparison of MarkovType with RB methods using the following metrics: \emph{number of parameters}, \emph{ITR per selection}, \emph{accuracy}, \emph{mean number of sequences per selection ($n_{\tau}$)}, and \emph{ITR per sequence}. We observe that all MarkovType models perform better than RB models in terms of ITR per selection and accuracy. However, RB methods have a lower $n_{\tau}$, indicating they make decisions faster than MarkovType methods, but with lower performance in terms of ITR per selection and accuracy. This highlights a trade-off between speed and accuracy. 

To analyze ITR with the time spent on decisions, we incorporate the number of sequences into the ITR calculation (ITR per sequence). We observe that despite requiring a higher $n_{\tau}$, all MarkovType models achieve better overall performance. When comparing the effect of discount factors, we observe that going from $\frac{2N-n-1}{N}$ to $\frac{1}{n^2}$ results in a decrease in $n_{\tau}$ and an increase in ITR per sequence. However, this trend does not continue when moving to $\frac{1}{n^3}$. As explained in Section~\ref{sec:dataset and model details}, the discount factors vary in the importance they assign to early correct classifications; i.e., the importance increases in the given order from $\frac{2N-n-1}{N}$ to $\frac{1}{n^3}$.

\begin{table*}[t!]
\centering
\caption{Number of parameters, ITR per selection, accuracy, mean number of sequences per selection ($n_{\tau}$), and ITR per sequence are evaluated with the threshold for MarkovType using different discount functions $d(\cdot)$, and RB - 1D and 2D. Note that classification sequences terminate when any posterior probability exceeds the threshold or when the maximum number of sequences (in this case, 10) is reached. Therefore, lower sequences per symbol with high accuracy are desired for good typing performance. We observe a trade-off between accuracy and the number of sequences per symbol. Even though RB methods perform quicker classification, MarkovType methods still achieve better ITR per sequence, where the mean number of sequences per symbol is included in the ITR calculation.}
\label{tab:results}
\begin{tabular}{|c|c|c|c|c|c|c|}
\hline
\textbf{Model} &
    \textbf{\begin{tabular}[c]{@{}c@{}}$d(n)$\end{tabular}} &
  \textbf{\begin{tabular}[c]{@{}c@{}}Num.\\ Par\end{tabular}} &
  \textbf{\begin{tabular}[c]{@{}c@{}}ITR w/$\tau$\\ per Selection $(\uparrow)$ \end{tabular}} &
  \textbf{\begin{tabular}[c]{@{}c@{}}Accuracy\\ w/$\tau$ $(\uparrow)$\end{tabular}} &
  \textbf{\begin{tabular}[c]{@{}c@{}}Num. Sequences\\ per Selection ($\downarrow$)\end{tabular}}
  &
  \textbf{\begin{tabular}[c]{@{}c@{}}ITR w/$\tau$\\ per Sequence $(\uparrow)$\end{tabular}}\\ \hline
\textit{MarkovType} &$\frac{2N-n\!-\!1}{N}$ & 479293  & \textbf{3.635$\pm$0.031} & \textbf{0.870$\pm$0.004}  & 4.956$\pm$0.183  & 0.734$\pm$0.023 \\ \hline
\textit{MarkovType}  &$\frac{1}{n}$ & 479293 & 3.550$\pm$0.110  & 0.859$\pm$0.015  &  4.832$\pm$0.213  &  0.735$\pm$0.026\\ \hline
\textit{MarkovType}  &$\frac{1}{n^2}$ & 479293 & 3.450$\pm$0.143  & 0.845$\pm$0.020   & 4.524$\pm$0.243 &  \textbf{0.763$\pm$0.012}\\ \hline
\textit{MarkovType}  &$\frac{1}{n^3}$ & 479293 & 3.502$\pm$0.074 & 
0.852$\pm$0.010  & 4.638$\pm$0.142  &  0.756$\pm$0.031\\ \hline
\textit{RB - 1D CNN}     & - & 542210  & 1.230$\pm$0.061  & 0.457$\pm$0.013 & \textbf{2.321$\pm$0.064} &  0.531$\pm$0.037 \\ \hline
\textit{RB - 2D CNN}    & - & 1603042 & 1.377$\pm$0.058  & 0.489$\pm$0.012 &  2.564$\pm$0.040  &  0.537$\pm$0.020\\ \hline
\end{tabular}
\end{table*}

\begin{figure*}[t!]
    \centering
    
    \begin{subfigure}{0.3\textwidth}
        \centering
        \includegraphics[width=\textwidth]{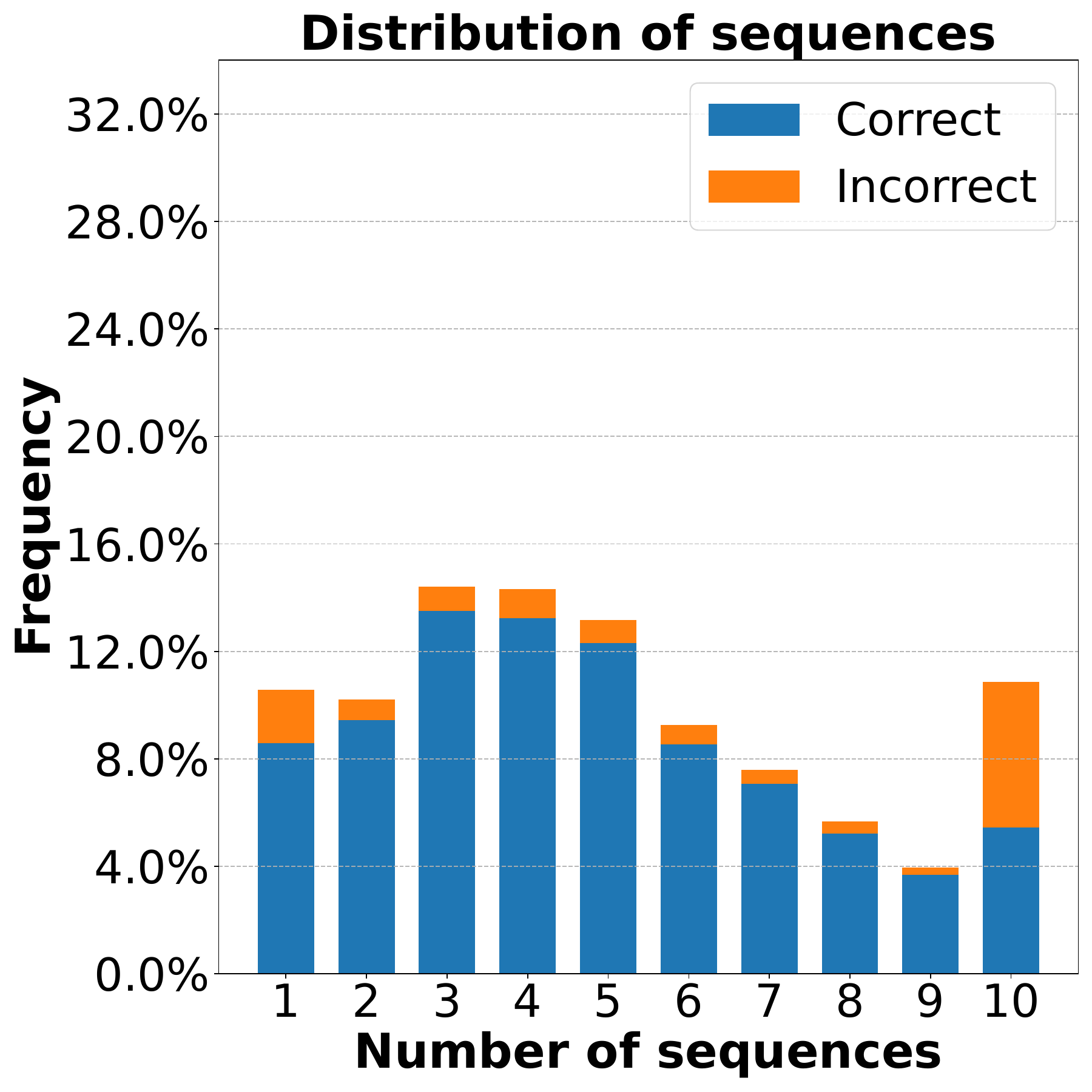}
        \subcaption{MarkovType $\left(\frac{2N-n\!-\!1}{N}\right)$}\label{fig:subfig1}
    \end{subfigure}
    \hfill
    \begin{subfigure}{0.3\textwidth}
        \centering
        \includegraphics[width=\textwidth]{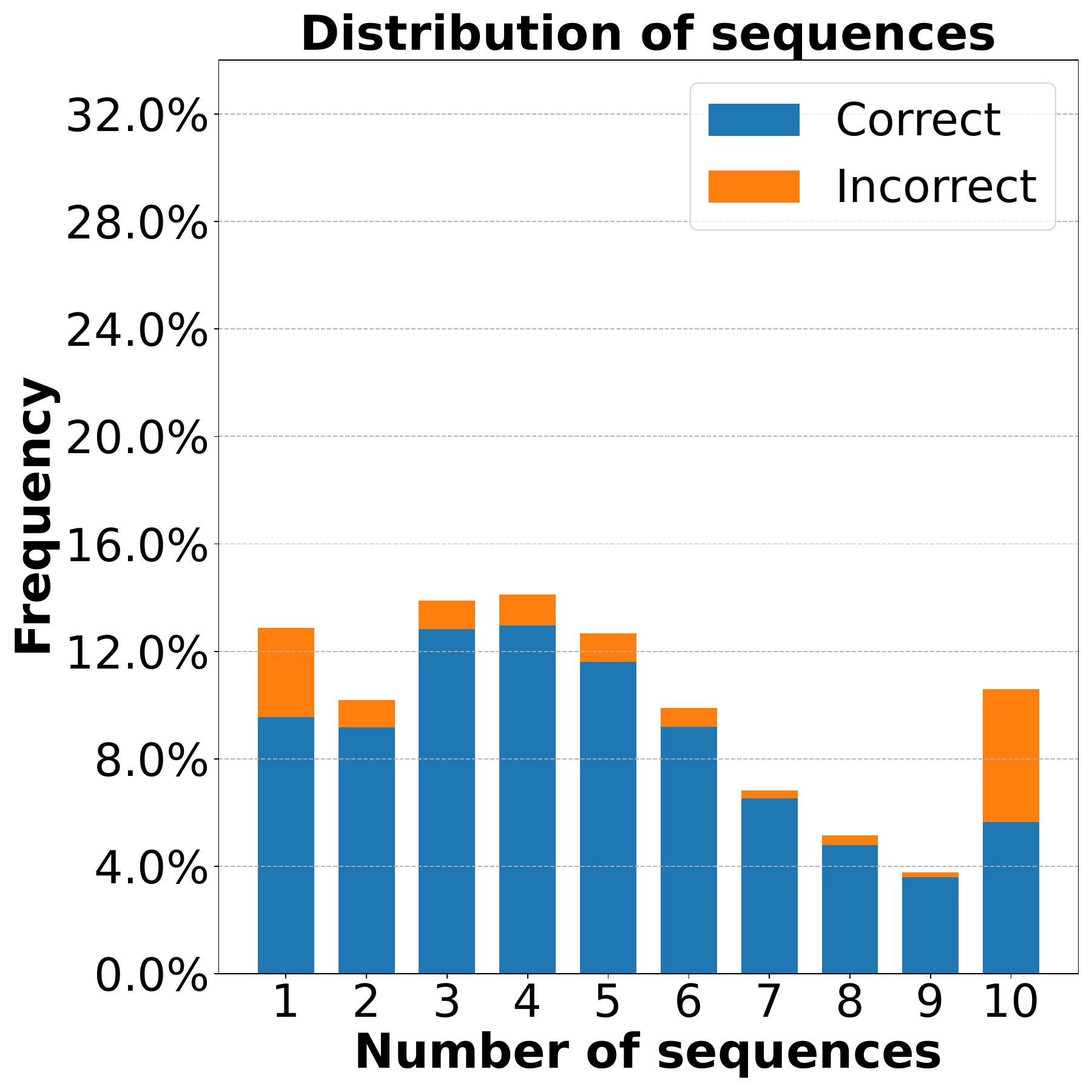}
        \subcaption{MarkovType $\left(\frac{1}{n}\right)$}\label{fig:subfig2}
    \end{subfigure}
    \hfill
    \begin{subfigure}{0.3\textwidth}
        \centering
        \includegraphics[width=\textwidth]{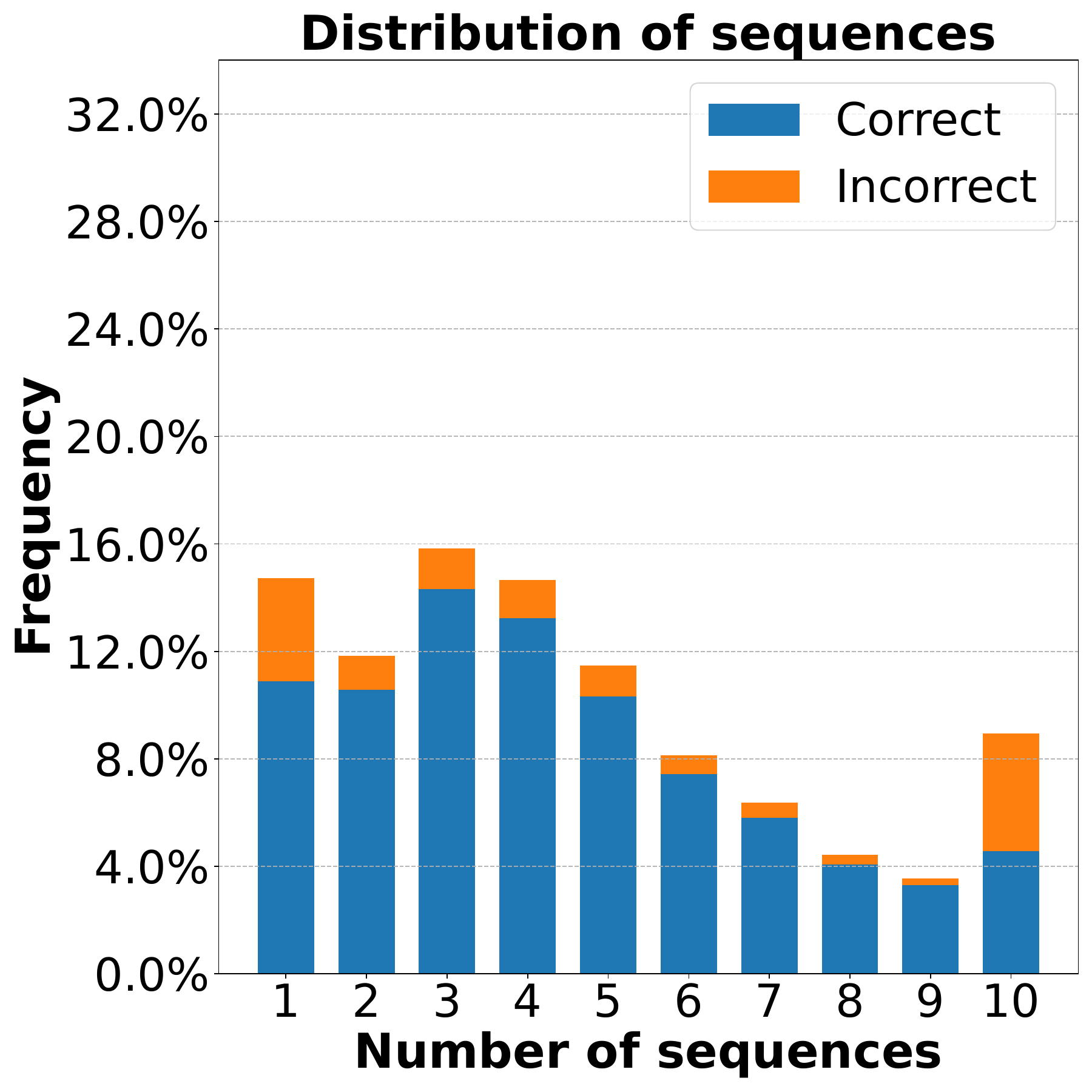}
        \subcaption{MarkovType $\left(\frac{1}{n^2}\right)$}\label{fig:subfig3}
    \end{subfigure}
    
    \vspace{1em}
    
    \begin{subfigure}{0.3\textwidth}
        \centering
        \includegraphics[width=\textwidth]{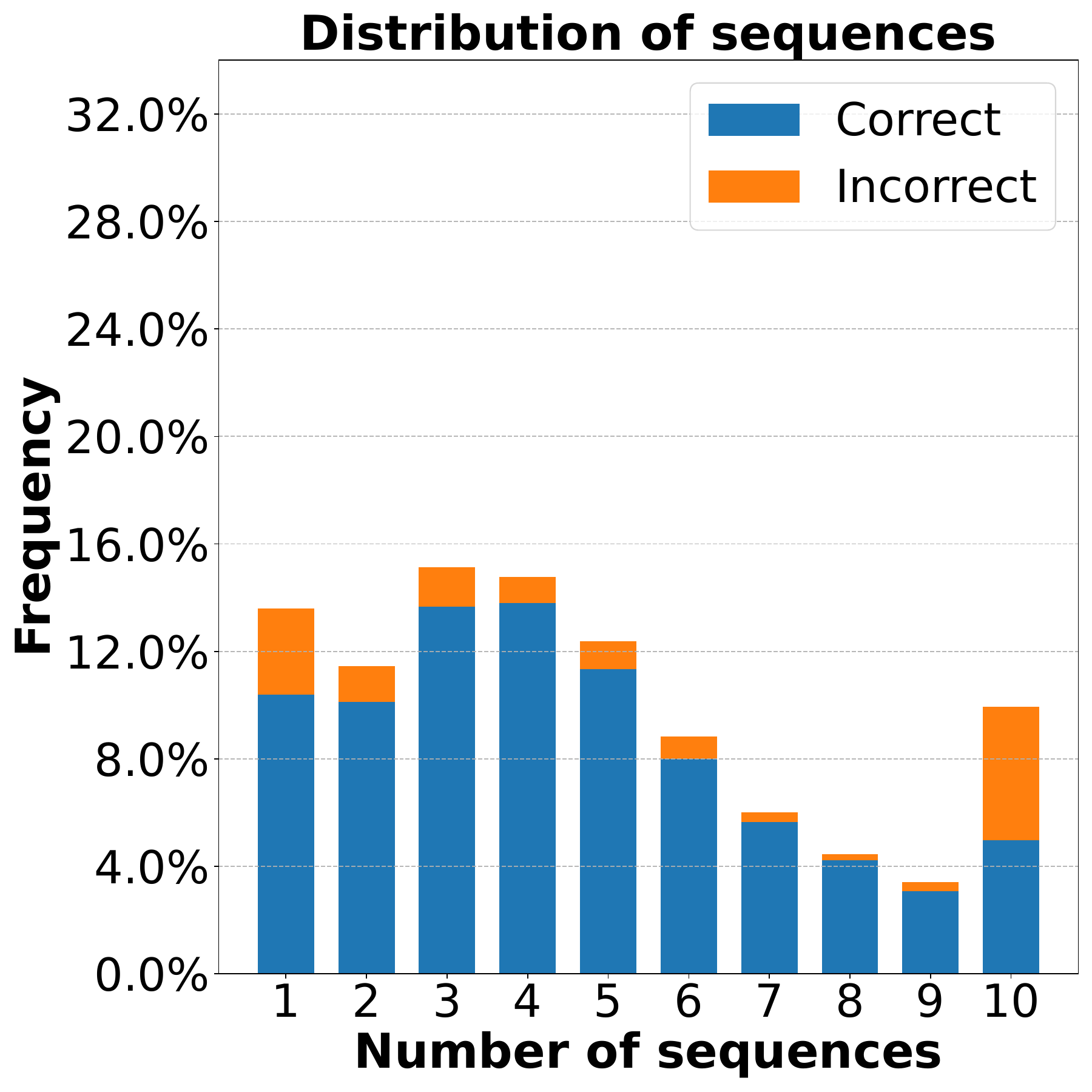}
        \subcaption{MarkovType $\left(\frac{1}{n^3}\right)$}\label{fig:subfig4}
    \end{subfigure}
    \hfill
    \begin{subfigure}{0.3\textwidth}
        \centering
        \includegraphics[width=\textwidth]{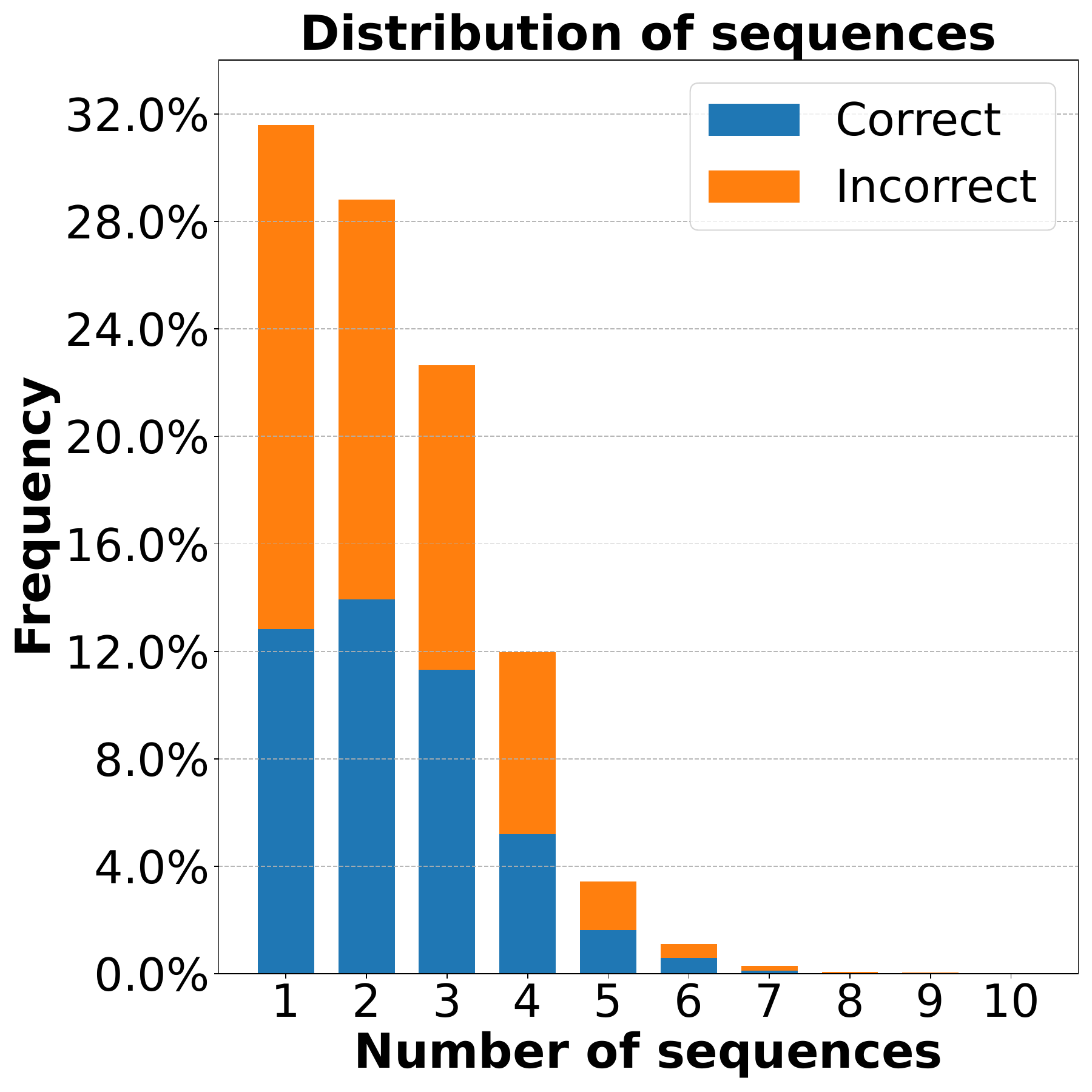}
        \subcaption{RB 1D - CNN}\label{fig:subfig5}
    \end{subfigure}
    \hfill
    \begin{subfigure}{0.3\textwidth}
        \centering
        \includegraphics[width=\textwidth]{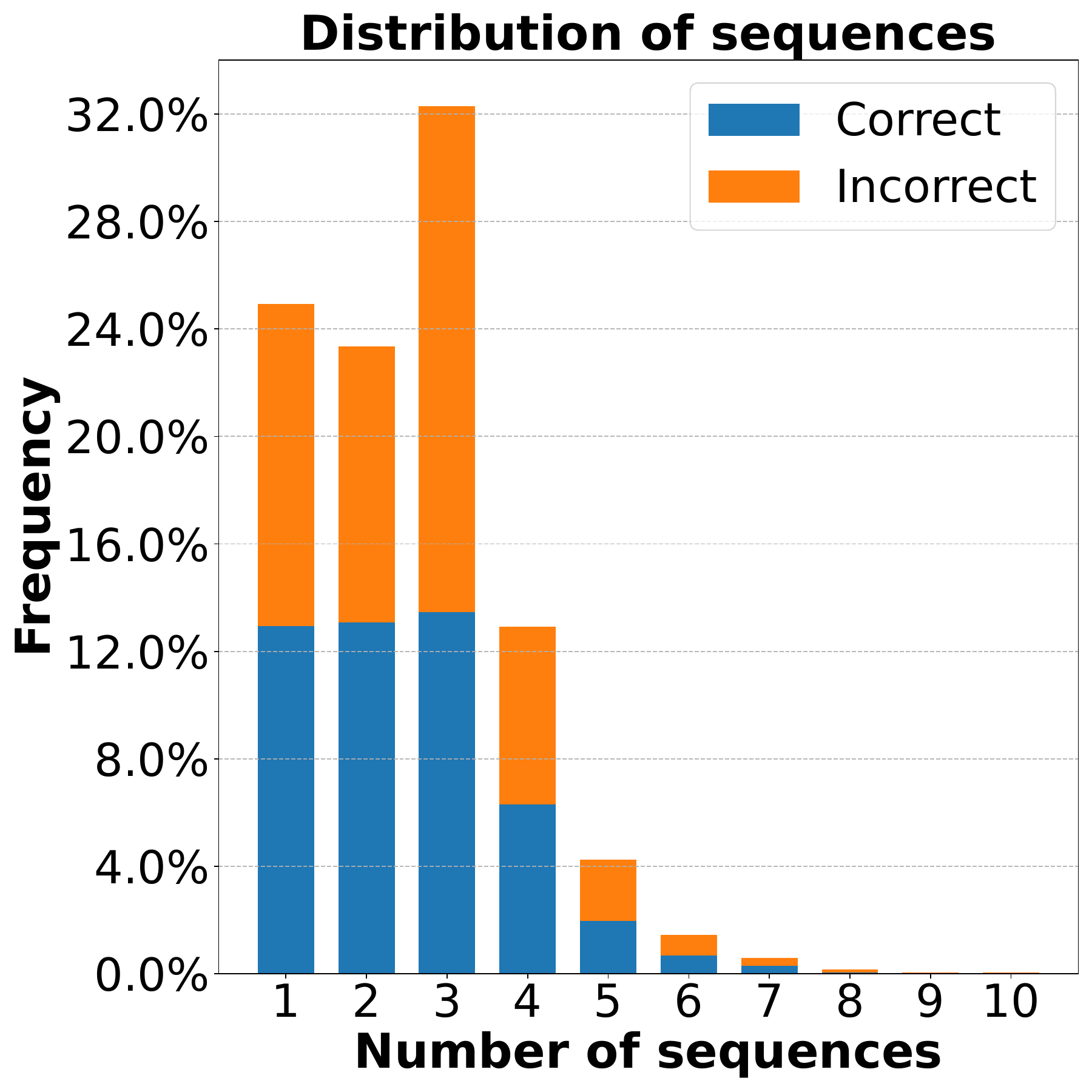}
        \subcaption{RB 2D - CNN}\label{fig:subfig6}
    \end{subfigure}
    
    \caption{Distribution of correctly and incorrectly classified symbols over 5 data splits across sequences (1-10). Note that classification sequences terminate when any posterior probability exceeds the threshold or when the maximum number of sequences (in this case, 10) is reached. At each sequence, the blue bar represents correct decisions, and the orange bar represents incorrect decisions. Together, they show the total number of decisions made at each sequence. We observe that although RB methods (e) and (f) make most of their decisions early on, these decisions are mostly incorrect. In contrast, MarkovType methods (a)-(d) tend to make decisions later than those using recursive Bayesian methods but achieve a higher ratio of correct decisions.}
    \label{fig:attempt_histogram}
\end{figure*}

Figure~\ref{fig:attempt_histogram} shows the distributions of correct and incorrect decisions made at each sequence. Similar to the observations in Table~\ref{tab:results}, RB methods make most of their decisions at earlier sequences and classify all 1000 symbols before sequence 8. MarkovType methods make decisions at later sequences, with some cases classified after sequence 7, but they achieve a higher accuracy rate at each sequence. We observe that going from $\frac{2N-n-1}{N}$ to $\frac{1}{n^2}$ results in more decisions being made in the first 4 sequences. However, this trend does not occur when moving from $\frac{1}{n^2}$ to $\frac{1}{n^3}$.

\begin{figure}[t!]
    \centering
    \includegraphics[width=\linewidth]{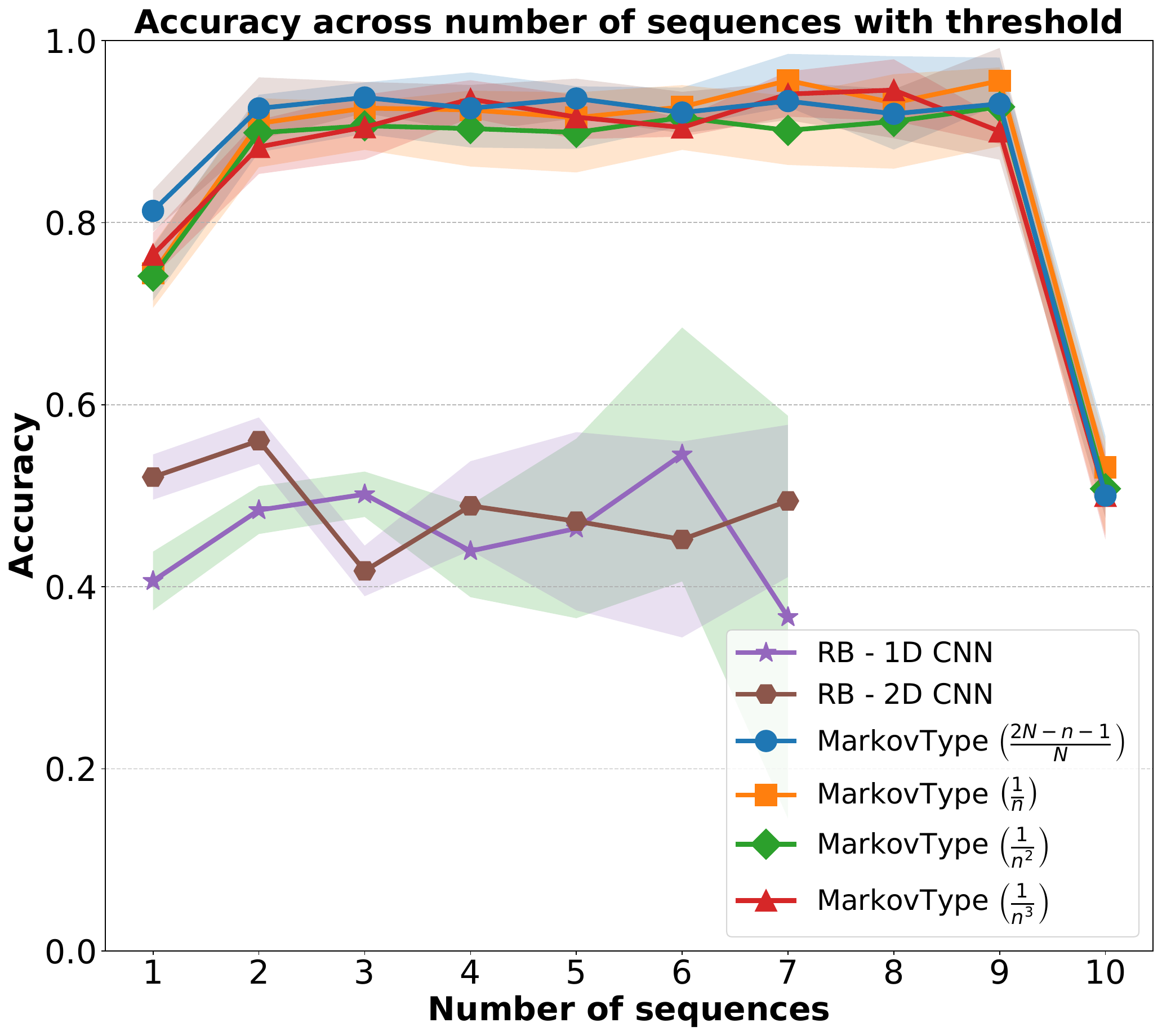}
    \caption{Accuracy across sequences with threshold $\tau$. Lines represent the mean accuracy and confidence bars indicate the standard deviation across the 5 data splits. This plot shows the accuracy of decisions made across sequences. Classification sequences terminate when any posterior probability exceeds the threshold or when the maximum number of sequences is reached. With RB methods, no symbols remain to be classified after the 7th sequence. MarkovType consistently achieves higher accuracy than RB methods.}  
    % \caption{Accuracy across sequences with threshold $\tau$. Lines represent the mean accuracy and confidence bars indicate the standard deviation across the 5 data splits. This plot shows the accuracy of decisions made across sequences. Note that classification sequences terminate when any posterior probability exceeds the threshold or when the maximum number of sequences is reached. With RB methods, no symbols remain to be classified after the 7th sequence. MarkovType consistently achieves higher accuracy than RB methods.} 
    \label{fig:results_acc_attempt_with_tau}
\end{figure}

To analyze the accuracy of decisions in more detail statistically, Figure~\ref{fig:results_acc_attempt_with_tau} examines the accuracy of decisions made at each sequence, showing performance over 5 data splits. The lines represent the mean accuracy, with confidence bars indicating the standard deviation across the 5 data splits. Figure~\ref{fig:results_acc_attempt_with_tau} clearly demonstrates the accuracy difference between MarkovType and RB methods, with MarkovType consistently achieving much higher accuracy than RB methods. In most sequences, the accuracy difference is about $0.4$.

\subsection{Without Threshold $\tau$}
\begin{figure}[t!]
    \centering
    \includegraphics[width=\linewidth]{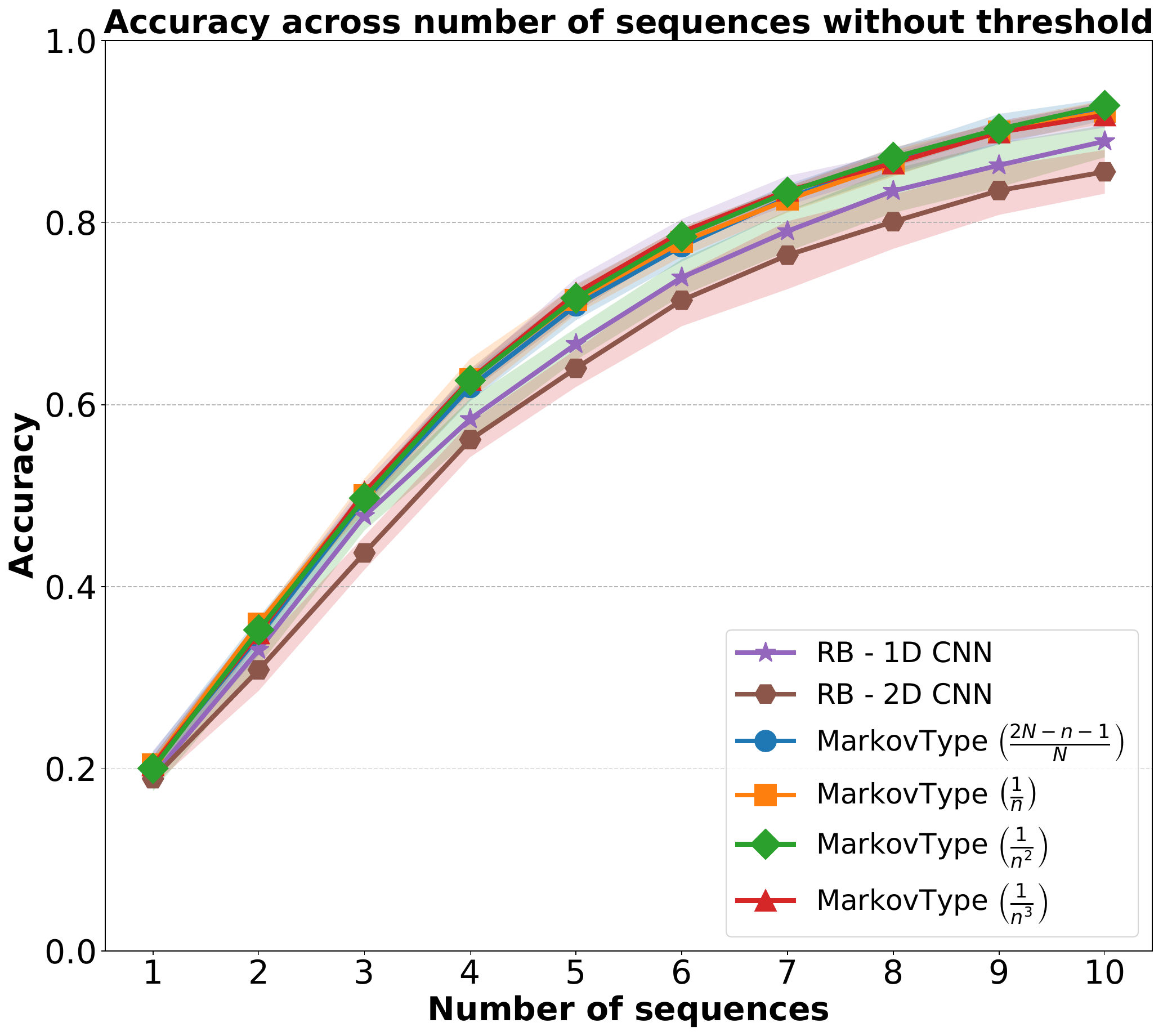}
    \caption{Accuracy across sequences without threshold $\tau$, with lines representing the mean accuracy and confidence bars indicating the standard deviation across the 5 data splits. This plot shows the accuracy of symbols with varying numbers of sequences from 1 to 10. Note that when the algorithm includes a threshold, early stopping occurs. Without early stopping, we can observe the upper limit of accuracy with different numbers of sequences. For instance, with a threshold, the model might make an incorrect prediction without an opportunity to correct it in subsequent sequences, resulting in differences in accuracy between cases with and without a threshold. At each sequence, the classification of 1000 symbols in the test set is performed. We see that MarkovType methods perform better at each sequence compared to RB methods, with the MarkovType method using the $\frac{1}{n^2}$ discount factor achieving the highest accuracy.}
    \label{fig:results_acc_n}
\end{figure}

In this section, we evaluate accuracy performance as a function of the number of sequences, ranging from 1 to 10. We present model performances without using the threshold $\tau$, with all 1000 symbols in the test set classified at each sequence and no early stopping. Without a threshold, the model can correct incorrect predictions in subsequent sequences, resulting in differences in accuracy between cases with and without a threshold. Figure~\ref{fig:results_acc_n} shows the accuracy of all symbols across different sequences, with the lines representing mean accuracy and confidence bars indicating the standard deviation across the 5 data splits. MarkovType methods have higher accuracy across all sequences, except the first, where all methods exhibit similar accuracy. By the 10th sequence, which demonstrates the limits of all methods, MarkovType achieves higher accuracy, with the method using the $\frac{1}{n^2}$ discount factor performing the best.

% \section{Conclusion}
\section{Conclusion}\label{sec:conclusion}
This work proposes a Markov Decision Process for non-invasive BCI typing systems (MarkovType) that formulates the RSVP typing task as a Partially Observed Markov Decision Process (POMDP) for recursive classification. To the best of our knowledge, this is the first work to formalize the RSVP typing task as a POMDP. Formulating the RSVP typing task as a POMDP is appropriate because the task involves multiple sequences where partial subsets of symbols from an alphabet are presented. We perform our analyses on a simulated typing task and compare them with Recursive Bayesian-based methods. Unlike MarkovType, they assume that responses to each query are independent and do not include the typing procedure during training. Our experiments show that MarkovType significantly outperforms Recursive Bayesian methods in terms of accuracy and information transfer rate per selection and symbol, with the symbol-based calculation accounting for the time spent on selection. However, we observe a trade-off between accuracy and speed, specifically the number of sequences required for a decision when a confidence threshold is used to terminate sequences. This opens the door for future improvements in achieving both fast and accurate classification. Additionally, all experiments are conducted on a large RSVP benchmark. 

Future work might focus on the real-world adaptation of MarkovType with limited data and user-specific training. In this work, we aim to keep MarkovType simple yet effective to facilitate its potential real-world use, with fewer parameters than the baselines. If the model proves too complex for real-world applications with limited data, simplifying the architecture could be considered. Our method is also trained globally in this work. For user-dependent cases, transfer learning approaches should be considered to adapt the model to different users in real-time.

\section{Acknowledgments}
This work was supported by NIH grant R01DC009834 as part of CAMBI. We thank all the members of CAMBI for their insightful discussions and valuable contributions.
\bibliography{ref}
\end{document}